\documentclass[runningheads]{llncs}

\usepackage{eccv}

\usepackage{eccvabbrv}

\usepackage{graphicx}
\usepackage{booktabs}

\usepackage[accsupp]{axessibility}  

\usepackage[hidelinks]{hyperref}

\usepackage{multirow}
\usepackage{colortbl}
\usepackage{xcolor}

\usepackage{subcaption}
\usepackage{amsmath}
\usepackage{amssymb}
\usepackage{amsfonts}
\usepackage{float}
\usepackage{makecell}
\usepackage{array}
\usepackage{pifont}
\usepackage{marvosym}

\begin{document}

\title{\texorpdfstring{ME-IQA: Memory-Enhanced \\Image Quality Assessment via Re-Ranking}{ME-IQA: Memory-Enhanced Image Quality Assessment via Re-Ranking}} 

\titlerunning{ME-IQA}

\author{Kanglong~Fan\inst{1}  \and
Tianhe~Wu\inst{1}\textsuperscript{(\Letter)} \and
Wen~Wen\inst{1} \and
Jianzhao~Liu\inst{2} \and \\
Le~Yang\inst{2} \and
Yabin~Zhang\inst{2} \and
Yiting~Liao\inst{2} \and
Junlin~Li\inst{2} \and
Li~Zhang\inst{2}}

\authorrunning{Fan et al.}

\institute{City University of Hong Kong, Hong Kong SAR, China
\and ByteDance Inc., China\\
\email{tianhewu-c@my.cityu.edu.hk}}

\maketitle

\begin{abstract}
 Reasoning‑induced vision-language models (VLMs) advance image quality assessment (IQA) with textual reasoning, yet their scalar scores often lack sensitivity and collapse to a few values, so-called \emph{discrete collapse}. We introduce ME‑IQA, a plug-and-play, test‑time memory-enhanced re-ranking framework. It (i) builds a memory bank and retrieves semantically and perceptually aligned neighbors using reasoning summaries, (ii) reframes the VLM as a probabilistic comparator to obtain pairwise preference probabilities and fuse this ordinal evidence with the initial score under Thurstone's Case V model, and (iii) performs gated reflection and consolidates memory to improve future decisions. This yields denser, distortion‑sensitive predictions and mitigates \emph{discrete collapse}. Experiments across multiple IQA benchmarks show consistent gains over strong reasoning-induced VLM baselines, existing non-reasoning IQA methods, and test‑time scaling alternatives.
  \keywords{IQA \and Memory \and Test Time Scaling}
\end{abstract}

\section{Introduction}
\label{sec:intro}

Image quality assessment (IQA), a fundamental problem in computer vision~\cite{wang2004image,wang2006modern}, underpins applications from mobile photography~\cite{fang2020perceptual} and video streaming~\cite{chen2019qoe} to image restoration~\cite{gu2020pipal}. The rapid progress of vision-language models (VLMs) has catalyzed a shift from conventional score regressors~\cite{zhang2021uncertainty,yang2022maniqa,zhang2023blind} toward VLM-based IQA~\cite{wu2024qalign,you2025teaching}, aiming to leverage perception-reasoning capabilities for more human-aligned judgments.

\begin{figure}[t]
    \centering
    \begin{subfigure}[t]{0.48\textwidth}
        \centering
        \includegraphics[width=0.9\linewidth]{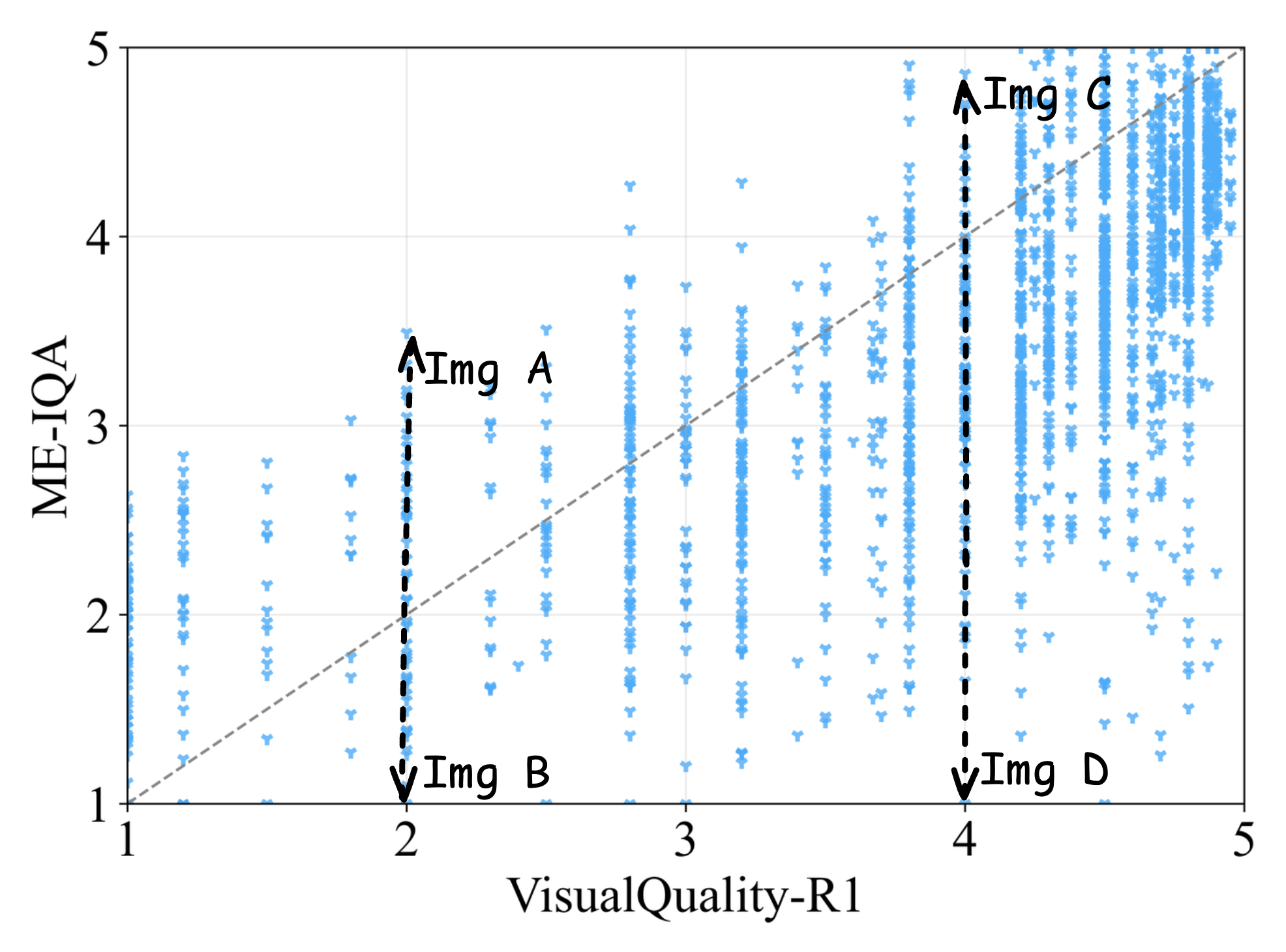}
        \caption{}
        \label{fig:dc_demo_a}
    \end{subfigure}
    \hfill
    \begin{subfigure}[t]{0.48\textwidth}
        \centering
        \includegraphics[width=0.85\linewidth]{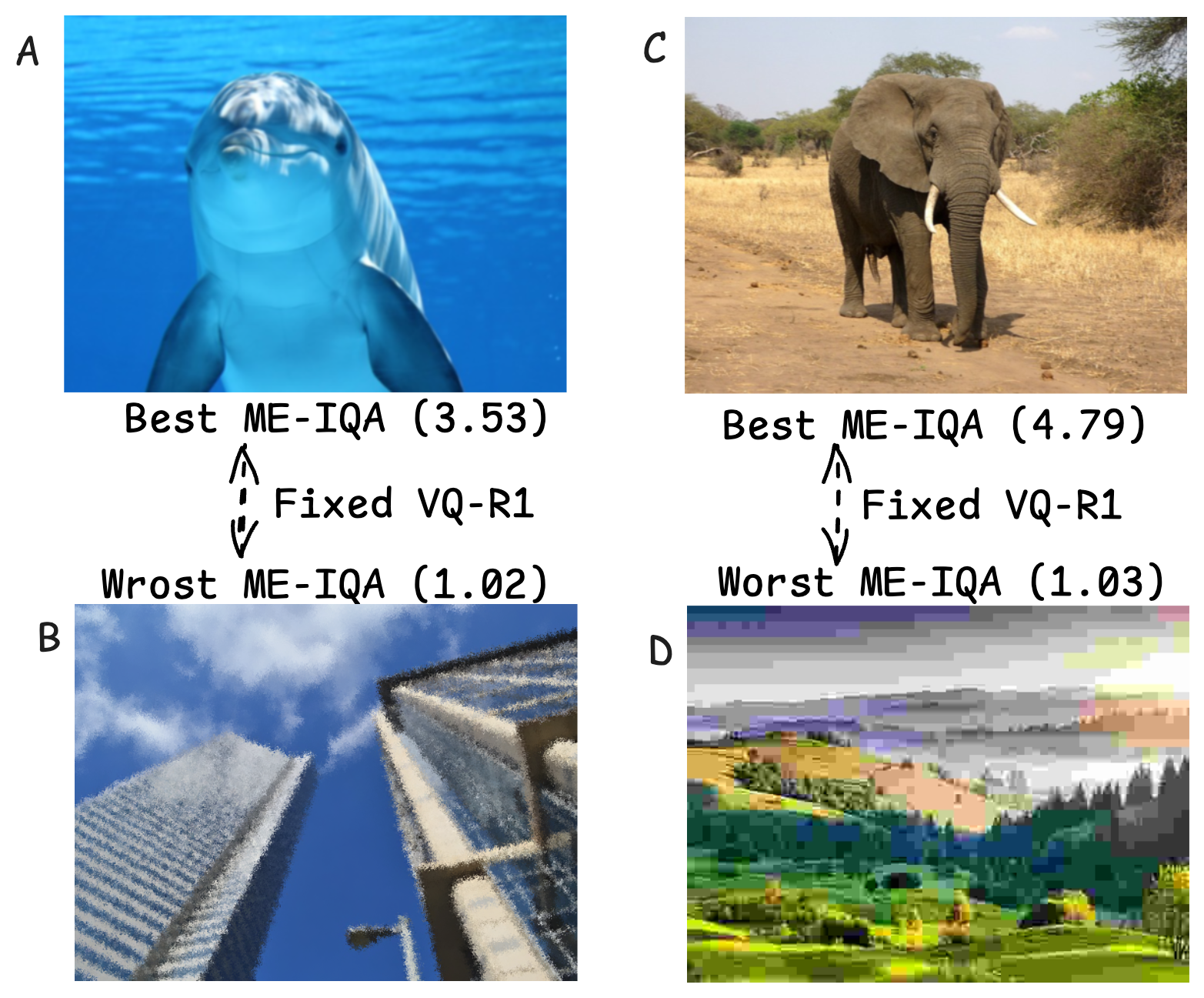}
        \caption{}
        \label{fig:dc_demo_b}
    \end{subfigure}
    \caption{ME-IQA mitigates discrete collapse and improves distortion sensitivity. (a) Scatter plot of ME-IQA versus baseline (VisualQuality-R1) scores on KADID-10K~\cite{lin2019kadid}. Points concentrate at a few discrete levels along the x-axis for the baseline, indicating \textit{discrete collapse}, whereas ME-IQA spreads predictions more densely. (b) Four highlighted cases illustrate substantial re-ranking: for a low baseline score around 2.0 (Img A vs. Img B) and a high baseline score around 4.0 (Img C vs. Img D), ME-IQA respectively elevates the higher-quality image and suppresses the lower-quality one relative to the baseline. }
    \label{fig:dc_demo}
   
\end{figure}

Reasoning-induced approaches~\cite{li2025qinsight,wu2025visualquality} prompt VLMs to generate step-by-step reasoning before emitting a scalar score and often generalize better than direct regression. Yet they frequently suffer from \emph{discrete collapse}: distinct-quality images receive coarse, nearly identical scores. As illustrated in Fig.~\ref{fig:dc_demo} (a), on KADID~\cite{lin2019kadid}, VisualQuality-R1~\cite{wu2025visualquality} produces scores concentrated at a few discrete levels, even though there are substantial perceptual differences among images assigned to the same level (see Fig.~\ref{fig:dc_demo} (b)). We attribute this to an objective mismatch: VLMs are pretrained to generate discrete tokens~\cite{brown2020language,devlin2019bert}, not continuous perceptual quantities~\cite{yang2022maniqa}. When forced into numeric prediction, they gravitate to textually salient numbers (\eg, ``3.0'', ``4.0'', ``5.0''), coarsely quantizing perception and dulling sensitivity to fine-grained distortions.

To resolve the collapse, existing work explores two potential remedies. The first extracts token probabilities for special numeric symbols and averages them to form a score~\cite{wu2024qalign,you2025teaching}. The second aggregates pairwise comparisons under Thurstone’s model~\cite{thurstone1927law}, as in Depict-QA~\cite{you2024depicting}. While both yield continuous outputs, each has drawbacks. Single-stimulus prompting lacks explicit comparative context and undershoots subtle quality differences. Pure pairwise schemes, though perceptually grounded, scale poorly on large sets (\eg, KonIQ-10K~\cite{hosu2020koniq}) and are impractical for online testing\footnote{A stream of queries arrives sequentially and each should be processed immediately without access to future inputs (``no peeking'').}.

Anchor-based designs strike a middle ground by comparing a query to a small fixed set (\eg, Compare2Score~\cite{zhu2024adaptive}), tapping the VLM’s comparative strength with manageable cost. However, static anchors under-represent long-tail or novel distortions and degrade under distribution shift. In contrast, human observers judge quality relative to contextually retrieved memories of similar stimuli~\cite{stewart2006decision,helson1964adaptation}, beyond static reference. A similar principle appears in retrieval-augmented generation (RAG)~\cite{lewis2020retrieval} and external memory for VLMs~\cite{maharana2024evaluating,hu2025evaluating}, where instance-specific context improves robustness.

Motivated by dynamic human perceptual memory, we propose \textbf{ME}-\textbf{IQA}, a test-time \textbf{M}emory-\textbf{E}nhanced re-ranking framework that mitigates \emph{discrete collapse} in reasoning-induced VLMs while avoiding the brittleness of static anchors. Concretely, for each query, ME-IQA (i) constructs a hybrid memory bank and retrieves a semantically and perceptually aligned neighborhood using reasoning-derived summaries as retrieval keys; (ii) reframes the VLM as a comparator to elicit pairwise preference probabilities and fuses this ordinal evidence with the initial score by optimizing Thurstone’s Case V~\cite{thurstone1927law}, yielding denser, perceptually sensitive predictions (see Fig.~\ref{fig:dc_demo}); and (iii) performs gated reflection and consolidates the processed case into memory, strengthening future decisions.

Because ME-IQA operates purely at test time, requires only black-box access to the underlying VLM, and leaves training untouched, it is \emph{plug-and-play}: practitioners can apply it to existing reasoning-induced VLMs without retraining, architectural changes, or extra supervision. Extensive experiments on standard IQA benchmarks show consistent gains over strong VLM baselines and test-time scaling alternatives, narrowing the gap to human MOS distributions while preserving the generalization benefits of reasoning-induced models. In summary, ME-IQA offers an efficient path toward fine-grained, human-aligned IQA, effectively countering \emph{discrete collapse} and improving sensitivity to nuanced distortions.

\section{Related Work}
\label{sec:related}

\subsection{NR-IQA Models}
\label{sec:rw:nriqa}

No-reference image quality assessment (NR-IQA) aims to predict visual quality from a single distorted image without access to any pristine reference. Early NR-IQA approaches rely on natural scene statistics and degradation-specific priors~\cite{mittal2012no,mittal2012making,wang2000blind,wang2003local,liu2009no,wang2011reduced}, modeling regularities of undistorted images to detect departures due to artifacts. With deep learning, convolutional~\cite{zhang2021uncertainty} and Transformer-based~\cite{yang2022maniqa} architectures became dominant, extracting semantics-aware features that better correlate with perceptual quality. Deep NR-IQA has been optimized either by regression loss~\cite{kang2014convolutional,ma2017end,bosse2017deep,talebi2018nima,ke2021musiq,yang2022maniqa,you2025teaching}, which treats quality as an absolute scalar, or by ranking loss~\cite{gao2015learning,burges2005learning,ma2017dipiq,thurstone1927law,zhang2021uncertainty,wang2021troubleshooting,wang2021active,zhang2022continual}, casting quality as an intrinsically relative quantity. In parallel, memory-like ideas appeared as exemplar storage and local interpolation~\cite{Li2011blind,ye2012no,xue2013learning,wu2015blind,wu2017blind,wang2023regression}, but were ultimately constrained by hand-crafted or shallow features.

Leveraging large-scale pretraining and multimodal priors, VLM-based NR-IQA has rapidly gained traction~\cite{wu2024comprehensive}. Representative supervised finetuning (SFT) methods include Q-Align~\cite{wu2024qalign} and DeQA-Score~\cite{you2025teaching}, which directly regress scores. More recently, reasoning-induced models generate textual reasoning before scoring~\cite{li2025qinsight,wu2025visualquality}: Q-Insight~\cite{li2025qinsight} and VisualQuality‑R1~\cite{wu2025visualquality} use reinforcement learning (RL) to promote explicit reasoning, and EvoQuality~\cite{wen2025evoquality} combines self-consistency voting with reinforcement learning to rank. Despite improved generalization, these models remain vulnerable to \emph{discrete collapse}, where scalar predictions concentrate on a few values, dulling sensitivity to fine-grained distortions.

\subsection{Memory for LLMs/VLMs}
\label{sec:rw:memory}

Beyond purely parametric memory, recent work equips LLMs/VLMs with explicit external memory and agentic modules~\cite{zhang2025survey}. Retrieval-augmented generation (RAG) conditions generation on context fetched at inference time~\cite{guu2020retrieval,lewis2020retrieval}; multimodal variants index images, captions, dense embeddings, or structured facts to ground reasoning and improve robustness under distribution shift~\cite{hu2023reveal,yasunaga2022retrieval}. In agentic settings, working memory may be represented as plain text~\cite{packer2023memgpt}, latent embeddings~\cite{wang2025m}, or structured graphs~\cite{xu2025mem,chhikara2025mem0}. Most existing work emphasizes personalization~\cite{zhang2025prime,zhong2024memorybank} and long-context management~\cite{maharana2024evaluating,wu2025longmemeval,hu2025evaluating}. A complementary line studies self-improving agents that learn from prior trajectories~\cite{ouyang2025reasoningbank,zhao2024expel,su2025learnbyinteract,zheng2024synapse,tang2025chemagent}.

Our framework specializes memory for perceptual quality assessment through (i) a hybrid memory bank balancing stability (offline anchors) and adaptability (online contrast cases), and (ii) reasoning-aware retrieval keyed by concise quality descriptions. Unlike prior VLM-based NR-IQA methods that rely solely on SFT or end-to-end RL, our method operates purely at test time and targets the \emph{discrete collapse} failure mode by injecting local ordinal constraints and consolidating hard cases for future retrieval.

\begin{figure}[!tbp]
\centering
\includegraphics[width=1\textwidth]{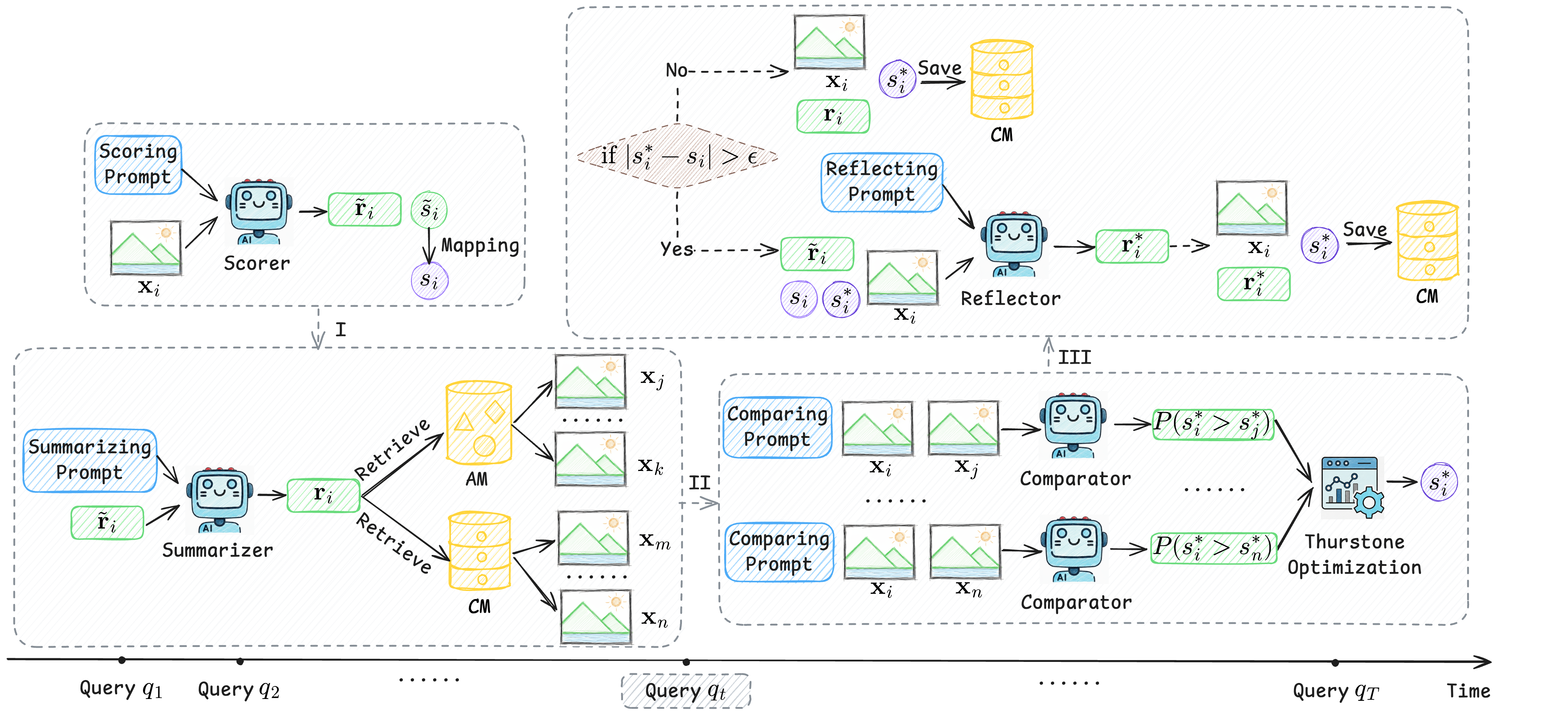}
\caption{Online testing process of ME-IQA. For each online query, the VLM produces a reasoning $\tilde{\mathbf{r}}_i$ and an initial score $\tilde{s}_i$, which is transformed to $s_i$ via a five-parameter monotonic mapping. The reasoning $\tilde{\mathbf{r}}_i$ is summarized as a concise quality description $\mathbf{r}_i$ and embedded to retrieve a neighborhood from a memory bank comprising Anchor Memory (AM) and Contrast Memory (CM). Acting as a comparator, the VLM estimates pairwise preference probabilities between the query and retrieved exemplars. The ordinal evidence is fused with the mapped initial score $s_i$ by optimizing Thurstone’s Case V to yield a refined score $s_i^*$. If $|s_i^*-s_i|>\epsilon$, a reflection step updates the description $\mathbf{r}_i$ to $\mathbf{r}_i^*$, and the case is consolidated to CM for better future decisions. 
}
\label{fig:ME-IQA_pipeline}
\end{figure}

\section{ME-IQA}
\label{sec:reiqa}

We adopt an online testing paradigm~\cite{wu2024streambench, wang2025inducing}, where a stream of NR-IQA queries $\mathcal{Q}=\{q_t\}_{t=1}^{T}$ arrives sequentially and should be processed on arrival, meaning no peeking into the future. This mirrors realistic streaming deployments that demand immediate, low-latency scoring and strict causality.

\noindent\textbf{Overview.} Given the query image $\mathbf{x}_i$ at time $t$ (omitting $t$ when clear), our system proceeds as in Fig.~\ref{fig:ME-IQA_pipeline}. A VLM first produces a free-form reasoning $\tilde{\mathbf{r}}_i$ and an initial scalar prediction $\tilde{s}_i$. We map $\tilde{s}_i$ to the target quality scale via a monotonic logistic transformation~\cite{sheikh2006statistical}, yielding $s_i$. We then summarize $\tilde{\mathbf{r}}_i$ into a concise, self-contained quality description $\mathbf{r}_i$ and embed it with a text encoder to retrieve a small neighborhood $\mathcal{N}$ from a hybrid memory bank $\mathcal{M}=\mathcal{M}_{\text{A}}\cup\mathcal{M}_{\text{C}}$ composed of (i) Anchor Memory (AM) and (ii) Contrast Memory (CM). Next, the VLM acts as a comparator to estimate pairwise preference probabilities between the query and retrieved exemplars, and the mapped initial score $s_i$ is fused with the ordinal evidence via a Thurstone Case~V objective to obtain a refined prediction $s_i^*$. A reflection mechanism is triggered upon inconsistency: if $|s_i^* - s_i|>\epsilon$, the VLM revises the description to $\mathbf{r}_i^*$ and the system consolidates the case into CM. Prompt templates are provided in the supplementary material. We now detail the memory bank (Sec.~\ref{sec:memorybank}) and quality re-ranking (Sec.~\ref{sec:rerank}).

\subsection{Memory Bank}
\label{sec:memorybank}

\noindent\textbf{Design.} To enable fine-grained, context-aware refinement at test time, we build a hybrid memory that balances stability and adaptability. AM is constructed offline from labeled anchors with ground-truth (GT) scores and provides a stable scaffold across the full quality range, mitigating drift and rectifying extreme mispredictions. CM is grown online from recently processed queries after re-ranking/reflection, continuously absorbing hard or underrepresented cases (distribution shifts, corner cases, emerging artifacts) to enhance local discrimination.

\noindent\textbf{Schema.} Each memory item $\mathbf{m}_i\in\mathcal{M}$ is a structured triple:
\begin{equation}
\mathbf{m}_i = \big(\mathbf{x}_i,\ \mathbf{r}_i \ \text{or}\ \mathbf{r}_i^*,\ s_i^*\big),
\end{equation}
where $\mathbf{r}_i$/$\mathbf{r}_i^*$ summarizes semantic content, clarity, texture detail, and perceptual distortions and their impact on quality; $s_i^*$ stores the GT for AM or the refined estimate for CM.

\noindent\textbf{Retrieval.} We employ reasoning-aware retrieval: the VLM first compresses the raw chain-of-thought $\tilde{\mathbf{r}}_i$ into a compact, quality-centric description $\mathbf{r}_i$, suppressing boilerplate and content-heavy detours (\eg, “let’s tackle this step by step”). Encoding $\mathbf{r}_i$ with a text encoder and using cosine similarity, we jointly search $\mathcal{M}_\text{A}$ and $\mathcal{M}_\text{C}$ under a fixed budget $K$. We split the budget uniformly:
\begin{equation}
K_{\text{A}}=\left\lfloor \frac{K}{2}\right\rfloor,\quad K_{\text{C}}=K - K_{\text{A}}.
\end{equation}
In AM, we perform GT-stratified retrieval by partitioning scores into $B$ bins over $[1,5]$ and taking $k_{\text{A}}=\lfloor K_{\text{A}}/B\rfloor$ nearest neighbors per bin (distributing any remainder randomly for coverage). In CM, we simply take the top-$K_{\text{C}}$ nearest neighbors. The final neighborhood is $\mathcal{N}=\mathcal{N}_{\text{A}}\cup\mathcal{N}_{\text{C}}$ with $|\mathcal{N}|=K$, used by the re-ranking module.

\noindent\textbf{Anchor Memory consolidation.} We pre-build $\mathcal{M}_{\text{A}}$ from a labeled set $\mathcal{D}=\{(\mathbf{x}_i, g_i)\}_{i=1}^{D}$ (\eg, KONIQ-10K). For each $\mathbf{x}_i$, the VLM emits $(\tilde{\mathbf{r}}_i,\tilde{s}_i)$. We fit the standard five-parameter logistic mapping~\cite{sheikh2006statistical} from $\{\tilde{s}_i\}$ to $\{g_i\}$ and store its parameters $\{\beta_k\}_{k=1}^{5}$. The mapped score is
\begin{equation}
\label{eq:five_param}
s_i = \beta_1\!\left(\frac{1}{2} - \frac{1}{1+\exp\big(\beta_2(\tilde{s}_i-\beta_3)\big)}\right) + \beta_4\,\tilde{s}_i + \beta_5,
\end{equation}
which aligns predictions to the GT scale $[1,5]$ (higher is better). If $|g_i-s_i|>\epsilon$, we prompt the VLM to reflect given $(\mathbf{x}_i,\tilde{\mathbf{r}}_i,s_i,g_i)$ and produce $\mathbf{r}_i^*$; otherwise, $\tilde{\mathbf{r}}_i$ is directly summarized to $\mathbf{r}_i$. We store $(\mathbf{x}_i,\mathbf{r}_i\ \text{or}\ \mathbf{r}_i^*,\ s_i^*=g_i)$. AM is static and persists across sessions, together with the learned mapping in Eq.~\eqref{eq:five_param}.

\noindent\textbf{Contrast Memory consolidation and pruning.} During online testing, after re-ranking the current query $q_t$, we add it to CM. Let $i=D+t$ index the current case. If $|s_i^*-s_i|>\epsilon$, we trigger reflection with $(\mathbf{x}_i,\tilde{\mathbf{r}}_i,s_i,s_i^*)$ to obtain $\mathbf{r}_i^*$. We store $(\mathbf{x}_i,\mathbf{r}_i\ \text{or}\ \mathbf{r}_i^*,\ s_i^*)$ with a cached text embedding $\mathbf{e}_i$ and count $c_i{=}1$. When $|\mathcal{M}_\text{C}|>C$, we apply agglomerative-style pruning~\cite{murtagh2012algorithms}: repeatedly find the closest pair by cosine distance $d(i,j)=1-\mathbf{e}_i^\top\mathbf{e}_j/(\|\mathbf{e}_i\|_2\|\mathbf{e}_j\|_2)$, keep the higher-count item as the prototype, remove the other item, and add the removed count to the survivor. Pruning stops at $|\mathcal{M}_\text{C}|\leq\alpha C$ ($\alpha{=}0.8$ by default).

\subsection{Quality Re-ranking}
\label{sec:rerank}

\noindent\textbf{Pairwise evidence from the VLM.} Given the query $\mathbf{x}_i$ and its neighborhood $\mathcal{N}$, the VLM serves as a probabilistic comparator. For each exemplar $j\in\mathcal{N}$, we prompt the model with a binary comparison (query is always “Image A”) and extract the probability of the token “A” from the answer distribution as the soft preference $y_{ij} = P(s_i^*>s_j^*)$~\cite{zhang2024generative,zhu2024adaptive}.

\noindent\textbf{Thurstone's Case~V with a weak prior.}
Let $s_j^*$ denote the stored score of neighbor $j$ (fixed constants from memory), and $\Phi(\cdot)$ the standard normal CDF. Under Thurstone’s Case~V model~\cite{thurstone1927law}, the model-implied probability is
\begin{equation}
p_{ij}=\Phi\!\left(\frac{s_i^*-s_j^*}{\sigma}\right),\qquad \sigma=1.
\end{equation}
We estimate $s_i^*$ by minimizing the binary cross-entropy (BCE) between $p_{ij}$ and $y_{ij}$ with a quadratic tether to the mapped initial score $s_i$:
\begin{equation}
\label{eq:objective}
\min_{s_i^*}\ \sum_{j\in\mathcal{N}} \mathrm{BCE}\big(p_{ij},y_{ij}\big)\ +\ \lambda\,(s_i^*-s_i)^2,
\end{equation}
where $\lambda$ controls the strength of the prior. The solution nudges $s_i^*$ toward ordinal evidence while remaining conservative when pairwise information is weak or ambiguous.

\noindent\textbf{Closed-form approximation.}
Although Eq.~\eqref{eq:objective} can be solved by gradient-based optimization, it does not admit a simple closed-form minimizer due to the probit likelihood in the BCE term.
For an efficient alternative, we use a common probit linearization that converts each soft preference into a pseudo-observation of the latent score.
Specifically, from the optimization objective $y_{ij}\approx \Phi(s_i^*-s_j^*)$ (with $\sigma=1$), we obtain
\begin{equation}
s_i^*\ \approx\ s_j^*+\Phi^{-1}(y_{ij})\ \triangleq\ \mu_{ij},
\end{equation}
For numerical stability, before applying $\Phi^{-1}$, we clip $y_{ij}$ into $[\delta,1-\delta]$ with a small $\delta>0$.
Replacing the BCE term in Eq.~\eqref{eq:objective} with a squared error yields a ridge-style objective
\begin{equation}
\min_{s_i^*}\ \sum_{j\in\mathcal{N}} \,(s_i^*-\mu_{ij})^2\ +\ \lambda\,(s_i^*-s_i)^2,
\end{equation}
whose closed-form solution is
\begin{equation}
s_i^*\ =\ \frac{\sum_{j\in\mathcal{N}} \,\mu_{ij}+\lambda s_i}{K+\lambda}.
\end{equation}
This approximation can be used for fast inference.

\begin{table}[!t]
\centering
\scriptsize
\caption{Performance improvement of ME-IQA over baseline VLM models and comparison with existing non-reasoning IQA methods across seven benchmarks. All models, except the closed-source ones, are trained on KONIQ using the same training protocol. The upper section shows PLCC results, and the lower section shows SRCC results. WAVG denotes the weighted average result, with weights proportional to the number of images in each dataset. The top two results are highlighted in \textbf{bold}
and \underline{underline}.}
\begin{tabular}{lcccccccc}
\toprule
\multicolumn{1}{l|}{\textbf{Method}}& \textbf{SPAQ} & \textbf{AGIQA}   & \textbf{LIVEW} & \textbf{KADID}  & \textbf{PIPAL} & \textbf{TID2013} & \multicolumn{1}{c|}{\textbf{CSIQ}}& \textbf{WAVG.}  \\
\midrule
\multicolumn{9}{l}{\cellcolor[HTML]{EFEFEF}\textit{Existing Non-reasoning Methods}} \\ 
\multicolumn{1}{l|}{NIQE}  & 0.664   & 0.533   &  0.449   & 0.430  & 0.120 & 0.110 & \multicolumn{1}{c|}{0.645}  &  0.349 \\
\multicolumn{1}{l|}{NIMA} & \multicolumn{1}{c}{0.838} & \multicolumn{1}{c}{0.715} & \multicolumn{1}{c}{0.814}    & \multicolumn{1}{c}{0.701}             & \multicolumn{1}{c}{0.536}            & \multicolumn{1}{c}{0.418}           & \multicolumn{1}{c|}{0.854}             & \multicolumn{1}{c}{0.641} \\
\multicolumn{1}{l|}{MUSIQ}&  0.868   &  0.722  &  0.789   &  0.668 & 0.567 & 0.257 & \multicolumn{1}{c|}{0.860}  &  0.621 \\ 
\multicolumn{1}{l|}{CLIP-IQA+}&  0.866  & 0.736   &  0.832   & 0.762  & 0.550  & 0.293 & \multicolumn{1}{c|}{\textbf{0.900}}  &  0.641 \\ 
\multicolumn{1}{l|}{MANIQA}&  0.768   &  0.723  & 0.849    & 0.526  & 0.603 & 0.172 & \multicolumn{1}{c|}{0.712}  &  0.582 \\ 
\multicolumn{1}{l|}{Q-Align}&  0.886  & 0.772   & 0.853  & \textbf{0.795} & 0.600  &0.282 & \multicolumn{1}{c|}{0.814}  &  0.663 \\
\multicolumn{1}{l|}{DeQA-Score}&   0.895  & 0.809   &  \textbf{0.891}   & 0.711  & 0.608 & 0.192 & \multicolumn{1}{c|}{\underline{0.868}}  & 0.652  \\
\multicolumn{1}{l|}{Compare2Score}&   0.873  & 0.756  &  0.857  & 0.720  & 0.596& 0.419 & \multicolumn{1}{c|}{0.761}  & 0.686  \\
\multicolumn{9}{l}{\cellcolor[HTML]{EFEFEF}\textit{Q-Insight}}\\
\multicolumn{1}{l|}{Baseline}& 0.901 & 0.823 & 0.859 & 0.732 & 0.604 & 0.477 & \multicolumn{1}{c|}{0.749} & 0.714 \\
\multicolumn{1}{l|}{ME-IQA}  & \underline{0.922} & 0.840 & 0.881 & 0.770 & \underline{0.635} & 0.520 & \multicolumn{1}{c|}{0.777} & \underline{0.744} \\
\multicolumn{9}{l}{\cellcolor[HTML]{EFEFEF}\textit{VisualQuality-R1}}   \\
\multicolumn{1}{l|}{Baseline}& 0.895 & 0.827 & 0.847 & 0.709 & 0.567 & 0.453 & \multicolumn{1}{c|}{0.750} & 0.698 \\
\multicolumn{1}{l|}{ME-IQA}  & 0.912 & \underline{0.850} & 0.868 & 0.741 & 0.599 & 0.490 & \multicolumn{1}{c|}{0.786} & 0.726 \\
\multicolumn{9}{l}{\cellcolor[HTML]{EFEFEF}\textit{EvoQuality}}   \\
\multicolumn{1}{l|}{Baseline}& 0.903 & 0.827 & 0.860 & 0.744 & 0.602 & 0.615 & \multicolumn{1}{c|}{0.821} & 0.748 \\
\multicolumn{1}{l|}{ME-IQA}  & \textbf{0.925} & 0.847 & \underline{0.887} & \underline{0.783} & \textbf{0.642} & \textbf{0.643} & \multicolumn{1}{c|}{0.851} & \textbf{0.777} \\
\multicolumn{9}{l}{\cellcolor[HTML]{EFEFEF}\textit{Doubao-Seed-1.6-vision}}   \\
\multicolumn{1}{l|}{Baseline}&    0.874   &    0.821   &    0.858   &   0.707    &   0.517    &    0.424   & \multicolumn{1}{c|}{0.729}      &  0.678     \\
\multicolumn{1}{l|}{ME-IQA}  &    0.888   &    \textbf{0.851}   &    0.872   &     0.740  &    0.561   &   0.469    & \multicolumn{1}{c|}{0.762}      &   0.711    \\
\multicolumn{9}{l}{\cellcolor[HTML]{EFEFEF}\textit{GPT-5}}   \\
\multicolumn{1}{l|}{Baseline}&    0.879   &    0.815   &  0.851     &  0.715     &   0.539    &   0.399    & \multicolumn{1}{c|}{0.701}      &    0.676   \\
\multicolumn{1}{l|}{ME-IQA}  &    0.891   &   0.847    &   0.860    &    0.739   &   0.566    &    0.447   & \multicolumn{1}{c|}{0.760}      &    0.706   \\
\toprule
\bottomrule
\multicolumn{9}{l}{\cellcolor[HTML]{EFEFEF}\textit{Existing Non-reasoning Methods}} \\ 
\multicolumn{1}{l|}{NIQE}& 0.664   & 0.533   &  0.449   & 0.430  & 0.120 & 0.110 & \multicolumn{1}{c|}{0.645}  &  0.349 \\

\multicolumn{1}{l|}{NIMA}   &  \multicolumn{1}{c}{0.856} & \multicolumn{1}{c}{0.654} & \multicolumn{1}{c}{0.771}               & \multicolumn{1}{c}{0.693}             & \multicolumn{1}{c}{0.486}            & \multicolumn{1}{c}{0.422}           & \multicolumn{1}{c|}{0.844}             & \multicolumn{1}{c}{0.616} \\

\multicolumn{1}{l|}{MUSIQ}&  0.863   & 0.630   &   0.830  & 0.650  & 0.535 & 0.248 & \multicolumn{1}{c|}{0.815}  & 0.592  \\ 
\multicolumn{1}{l|}{CLIP-IQA+}&  0.864   &  0.685  &  0.805   & 0.756  & 0.499 & 0.309 & \multicolumn{1}{c|}{\textbf{0.890}}  & 0.618  \\ 
\multicolumn{1}{l|}{MANIQA}&  0.758    &  0.636  &  0.832   & 0.488  & 0.551 & 0.116 & \multicolumn{1}{c|}{0.675}  &  0.534 \\ 

\multicolumn{1}{l|}{Q-Align}&  0.887    & 0.735   & 0.860  & \textbf{0.792} & 0.518 & 0.284 & \multicolumn{1}{c|}{0.750}  & 0.631  \\
\multicolumn{1}{l|}{DeQA-Score}&  0.895   & 0.729   &  \textbf{0.871}  &  0.716 & 0.565 & 0.132 & \multicolumn{1}{c|}{\underline{0.847}}  &  0.614 \\
\multicolumn{1}{l|}{Compare2Score}&   0.856 & 0.758  &  0.821  & 0.731  & \textbf{0.588}& 0.456 & \multicolumn{1}{c|}{0.771}  & 0.688  \\
\multicolumn{9}{l}{\cellcolor[HTML]{EFEFEF}\textit{Q-Insight}}\\
\multicolumn{1}{l|}{Baseline}& 0.899 & 0.766 & 0.817 & 0.737 & 0.528 & 0.463 & \multicolumn{1}{c|}{0.717} & 0.683 \\
\multicolumn{1}{l|}{ME-IQA}  & \textbf{0.924} & \underline{0.796} & 0.842 & \underline{0.785} & \underline{0.566} & \underline{0.508} & \multicolumn{1}{c|}{0.751} & \underline{0.719} \\
\multicolumn{9}{l}{\cellcolor[HTML]{EFEFEF}\textit{VisualQuality-R1}}   \\
\multicolumn{1}{l|}{Baseline}& 0.887 & 0.760 & 0.831 & 0.703 & 0.509 & 0.405 & \multicolumn{1}{c|}{0.697} & 0.661 \\
\multicolumn{1}{l|}{ME-IQA}  & 0.899 & 0.791 & 0.853 & 0.753 & 0.551 & 0.449 & \multicolumn{1}{c|}{0.737} & 0.696 \\
\multicolumn{9}{l}{\cellcolor[HTML]{EFEFEF}\textit{EvoQuality}}   \\
\multicolumn{1}{l|}{Baseline}& 0.901 & 0.782 & 0.829 & 0.738 & 0.543 & 0.581 & \multicolumn{1}{c|}{0.778} & 0.716 \\
\multicolumn{1}{l|}{ME-IQA}  & \underline{0.919} & \textbf{0.809} & \underline{0.860} & 0.783 & \textbf{0.588} & \textbf{0.619} & \multicolumn{1}{c|}{0.815} & \textbf{0.751} \\
\multicolumn{9}{l}{\cellcolor[HTML]{EFEFEF}\textit{Doubao-Seed-1.6-vision}}   \\
\multicolumn{1}{l|}{Baseline}&    0.867   &   0.748    &   0.837    &    0.709   &    0.483   &   0.387    & \multicolumn{1}{c|}{0.688}      &   0.648    \\
\multicolumn{1}{l|}{ME-IQA}  &    0.887   &    0.775   &    0.846   &   0.746    &    0.533   &    0.453   & \multicolumn{1}{c|}{0.739}      &    0.687  \\
\multicolumn{9}{l}{\cellcolor[HTML]{EFEFEF}\textit{GPT-5}}   \\
\multicolumn{1}{l|}{Baseline}&     0.873  &   0.737    &    0.823   &     0.701  &    0.496   &    0.373   & \multicolumn{1}{c|}{0.678}      &   0.644    \\
\multicolumn{1}{l|}{ME-IQA}  &  0.887     &    0.763   &    0.832   &    0.744   &   0.560    &      0.424 & \multicolumn{1}{c|}{0.736}      &  0.683    \\
\bottomrule
\end{tabular} 
\label{tab:compare_baseline}
\end{table}

\begin{table}[!t]
\centering
\caption{Comparison with test-time scaling alternatives across seven IQA benchmarks. All models are trained on KONIQ using the same training protocol. The upper section shows PLCC results and the lower section shows SRCC results.}
\resizebox{1.0\textwidth}{!}{
\begin{tabular}{lccccccccc}
\toprule

\multicolumn{1}{l|}{\textbf{Method}}& \multicolumn{1}{c|}{\textbf{Sec/Img}}&\textbf{SPAQ} & \textbf{AGIQA}   & \textbf{LIVEW} & \textbf{KADID}  & \textbf{PIPAL} & \textbf{TID2013} & \multicolumn{1}{c|}{\textbf{CSIQ}}& \textbf{WAVG.}  \\

\midrule
\multicolumn{10}{l}{\cellcolor[HTML]{EFEFEF}\textit{Q-Insight}}\\
\multicolumn{1}{l|}{Maj@64}    &    \multicolumn{1}{c|}{34.97}   & 0.912 & 0.835 & 0.870 & 0.740 & 0.611 & 0.487 & \multicolumn{1}{c|}{0.760} & 0.724 \\
\multicolumn{1}{l|}{Mean@64}   &  \multicolumn{1}{c|}{34.97}      & 0.915 & \textbf{0.842} & 0.875 & 0.738 & \textbf{0.615} & 0.490 & \multicolumn{1}{c|}{0.757} & 0.727 \\
\multicolumn{1}{l|}{Compare2Score@32} & \multicolumn{1}{c|}{14.02}& 0.887 & 0.801 & 0.842 & 0.715 & 0.589 & 0.460 & \multicolumn{1}{c|}{0.740} & 0.698 \\
\multicolumn{1}{l|}{ME-IQA@32}  &  \multicolumn{1}{c|}{14.34}    & \textbf{0.922} & 0.840 & \textbf{0.881} & \textbf{0.770} & 0.635 & \textbf{0.520} & \multicolumn{1}{c|}{\textbf{0.777}} & \textbf{0.744} \\
\hline
\multicolumn{10}{l}{\cellcolor[HTML]{EFEFEF}\textit{VisualQuality-R1}}   \\
\multicolumn{1}{l|}{Maj@64}   & \multicolumn{1}{c|}{34.92}       & 0.903 & 0.842 & 0.856 & 0.722 & 0.580 & 0.465 & \multicolumn{1}{c|}{0.762} & 0.710 \\
\multicolumn{1}{l|}{Mean@64}   &  \multicolumn{1}{c|}{34.92}     & 0.906 & 0.849 & 0.863 & 0.720 & 0.583 & 0.471 & \multicolumn{1}{c|}{0.772} & 0.714 \\
\multicolumn{1}{l|}{Compare2Score@32}& \multicolumn{1}{c|}{13.88}& 0.881 & 0.819 & 0.833 & 0.689 & 0.549 & 0.440 & \multicolumn{1}{c|}{0.737} & 0.684 \\
\multicolumn{1}{l|}{ME-IQA@32}  &  \multicolumn{1}{c|}{14.30}   & \textbf{0.912} & \textbf{0.850} & \textbf{0.868} & \textbf{0.741} & \textbf{0.599} & \textbf{0.490} & \multicolumn{1}{c|}{\textbf{0.786}} & \textbf{0.726} \\
\hline
\multicolumn{10}{l}{\cellcolor[HTML]{EFEFEF}\textit{EvoQuality}}   \\
\multicolumn{1}{l|}{Maj@64}   & \multicolumn{1}{c|}{34.95}     & 0.915 & 0.839 & 0.876 & 0.757 & 0.613 & 0.624 & \multicolumn{1}{c|}{0.832} & 0.759 \\
\multicolumn{1}{l|}{Mean@64}  & \multicolumn{1}{c|}{34.95}     & 0.922 & \textbf{0.847} & 0.883 & 0.749 & 0.610 & 0.623 & \multicolumn{1}{c|}{0.837} & 0.761 \\
\multicolumn{1}{l|}{Compare2Score@32}&\multicolumn{1}{c|}{13.93} & 0.888 & 0.815 & 0.853 & 0.725 & 0.590 & 0.602 & \multicolumn{1}{c|}{0.809} & 0.734 \\
\multicolumn{1}{l|}{ME-IQA@32}  &  \multicolumn{1}{c|}{14.32}   & \textbf{0.925} & \textbf{0.847} & \textbf{0.887} & \textbf{0.783} & \textbf{0.642} & \textbf{0.643} & \multicolumn{1}{c|}{\textbf{0.851}} & \textbf{0.777} \\
\toprule
\bottomrule
\multicolumn{10}{l}{\cellcolor[HTML]{EFEFEF}\textit{Q-Insight}}\\
\multicolumn{1}{l|}{Maj@64}    & \multicolumn{1}{c|}{34.97}       & 0.917 & 0.788 & 0.832 & 0.747 & 0.537 & 0.487 & \multicolumn{1}{c|}{0.733} & 0.700 \\
\multicolumn{1}{l|}{Mean@64}     & \multicolumn{1}{c|}{34.97}     & \textbf{0.924} & \textbf{0.796} & 0.838 & 0.740 & 0.532 & 0.485 & \multicolumn{1}{c|}{0.724} & 0.700 \\
\multicolumn{1}{l|}{Compare2Score@32}& \multicolumn{1}{c|}{14.02} & 0.890 & 0.756 & 0.799 & 0.711 & 0.511 & 0.449 & \multicolumn{1}{c|}{0.711} & 0.668 \\
\multicolumn{1}{l|}{ME-IQA@32}   & \multicolumn{1}{c|}{14.34}    & \textbf{0.924} & \textbf{0.796} & \textbf{0.842} & \textbf{0.785} & \textbf{0.566} & \textbf{0.508} & \multicolumn{1}{c|}{\textbf{0.751}} & \textbf{0.719} \\
\hline
\multicolumn{10}{l}{\cellcolor[HTML]{EFEFEF}\textit{VisualQuality-R1}}   \\
\multicolumn{1}{l|}{Maj@64}     & \multicolumn{1}{c|}{34.92}      & 0.905 & 0.779 & 0.851 & 0.725 & 0.534 & 0.422 & \multicolumn{1}{c|}{0.715} & 0.681 \\
\multicolumn{1}{l|}{Mean@64}     & \multicolumn{1}{c|}{34.92}     & \textbf{0.909} & 0.789 & \textbf{0.858} & 0.719 & 0.530 & 0.427 & \multicolumn{1}{c|}{0.719} & 0.683 \\
\multicolumn{1}{l|}{Compare2Score@32}& \multicolumn{1}{c|}{13.88} & 0.876 & 0.747 & 0.818 & 0.687 & 0.491 & 0.392 & \multicolumn{1}{c|}{0.683} & 0.647 \\
\multicolumn{1}{l|}{ME-IQA@32}    & \multicolumn{1}{c|}{14.30}   & 0.899 & \textbf{0.791} & 0.853 & \textbf{0.753} & \textbf{0.551} & \textbf{0.449} & \multicolumn{1}{c|}{\textbf{0.737}} & \textbf{0.696} \\
\hline
\multicolumn{10}{l}{\cellcolor[HTML]{EFEFEF}\textit{EvoQuality}}   \\
\multicolumn{1}{l|}{Maj@64}    & \multicolumn{1}{c|}{34.95}       & 0.919 & 0.799 & 0.850 & 0.753 & 0.560 & 0.597 & \multicolumn{1}{c|}{0.800} & 0.734 \\
\multicolumn{1}{l|}{Mean@64}   & \multicolumn{1}{c|}{34.95}       & \textbf{0.924} & \textbf{0.811} & \textbf{0.866} & 0.747 & 0.557 & 0.600 & \multicolumn{1}{c|}{0.812} & 0.738 \\
\multicolumn{1}{l|}{Compare2Score@32} & \multicolumn{1}{c|}{13.93}& 0.885 & 0.770 & 0.817 & 0.719 & 0.531 & 0.572 & \multicolumn{1}{c|}{0.760} & 0.703 \\
\multicolumn{1}{l|}{ME-IQA@32}   & \multicolumn{1}{c|}{14.32}    & 0.919 & 0.809 & 0.860 & \textbf{0.783} & \textbf{0.588} & \textbf{0.619} & \multicolumn{1}{c|}{\textbf{0.815}} & \textbf{0.751} \\
 \bottomrule
\end{tabular} 
}   
\label{tab:compare_scaling}
\end{table}

\section{Experiments}

\subsection{Experimental Setup}
\label{sec:exp:setup}

\noindent\textbf{Models under test.} We evaluate five reasoning-induced VLMs: Q-Insight~\cite{li2025qinsight}, VisualQuality‑R1~\cite{wu2025visualquality}, EvoQuality~\cite{wen2025evoquality}, and two proprietary frontier models, Doubao‑Seed‑1.6‑Vision and GPT‑5. Q‑Insight, VisualQuality‑R1, and EvoQuality are built on Qwen‑2.5‑VL~\cite{bai2025qwen2} and further fine‑tuned via GRPO~\cite{shao2024deepseekmath}, representing strong open‑source baselines for reasoning‑based IQA. Unless noted, we use their variants fine‑tuned on KONIQ‑10K, and we construct the anchor memory for all models uniformly from KONIQ‑10K to ensure fairness.

\noindent\textbf{Datasets.} We conduct zero‑shot evaluation on seven datasets spanning authentic, AI‑generated, and synthetic distortions following the same settings in~\cite{wen2025evoquality}: (i) authentic: SPAQ~\cite{fang2020perceptual}, LIVEW~\cite{ghadiyaram2015massive}; (ii) AI‑generated: AGIQA~\cite{li2023agiqa}; (iii) synthetic: CSIQ~\cite{larson2010most}, TID2013~\cite{ponomarenko2015image}, KADID~\cite{lin2019kadid}, and PIPAL~\cite{gu2020pipal}.

\noindent\textbf{Implementation details.} For the three open‑source models, reasoning traces are sufficiently concise; we therefore bypass the summarization step and directly encode the raw reasoning into retrieval embeddings. Quality descriptions are embedded with Qwen3‑Embedding‑0.6B~\cite{qwen3embedding}. Retrieval uses $K=32$ neighbors, split into $K_\text{A}=16$ from AM and $K_\text{C}=16$ from CM. AM is GT‑stratified into $B=5$ bins with $k_\text{A}=\lfloor 16/5 \rfloor=3$ items per bin and the remaining one item assigned to a random bin; CM contributes the top‑$K_\text{C}$ nearest neighbors (cosine similarity). The Thurstone fusion prior is set to $\lambda=0.01$ and the reflection gate to $\epsilon=0.75$.

\begin{figure*}[htbp]
  \centering
  \begin{subfigure}[b]{0.25\textwidth}
    \centering
    \includegraphics[width=\textwidth]{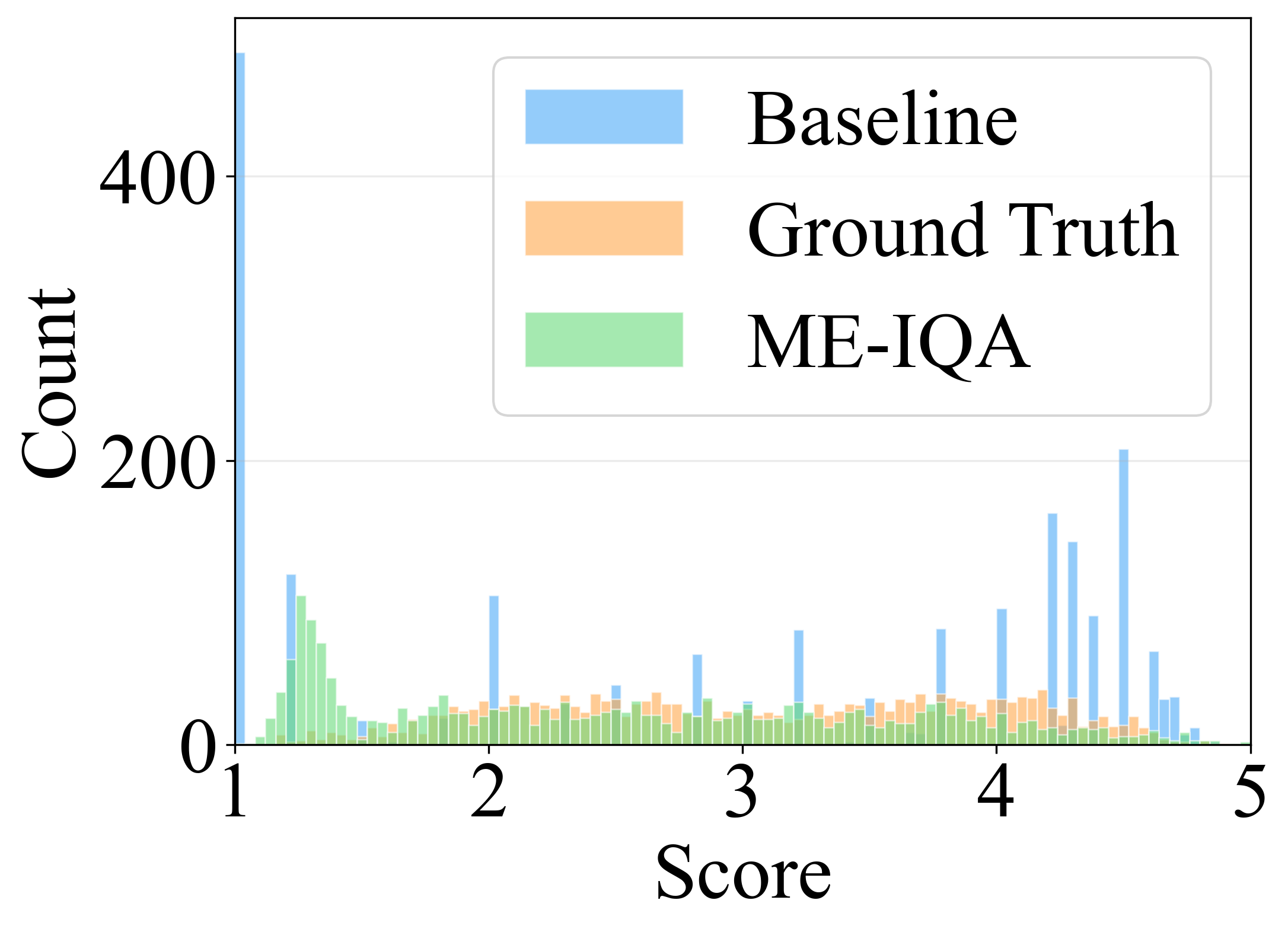}
    \subcaption{SPAQ}
    \label{fig:vqr1-spaq}
  \end{subfigure}\hfill
  \begin{subfigure}[b]{0.25\textwidth}
    \centering
    \includegraphics[width=\textwidth]{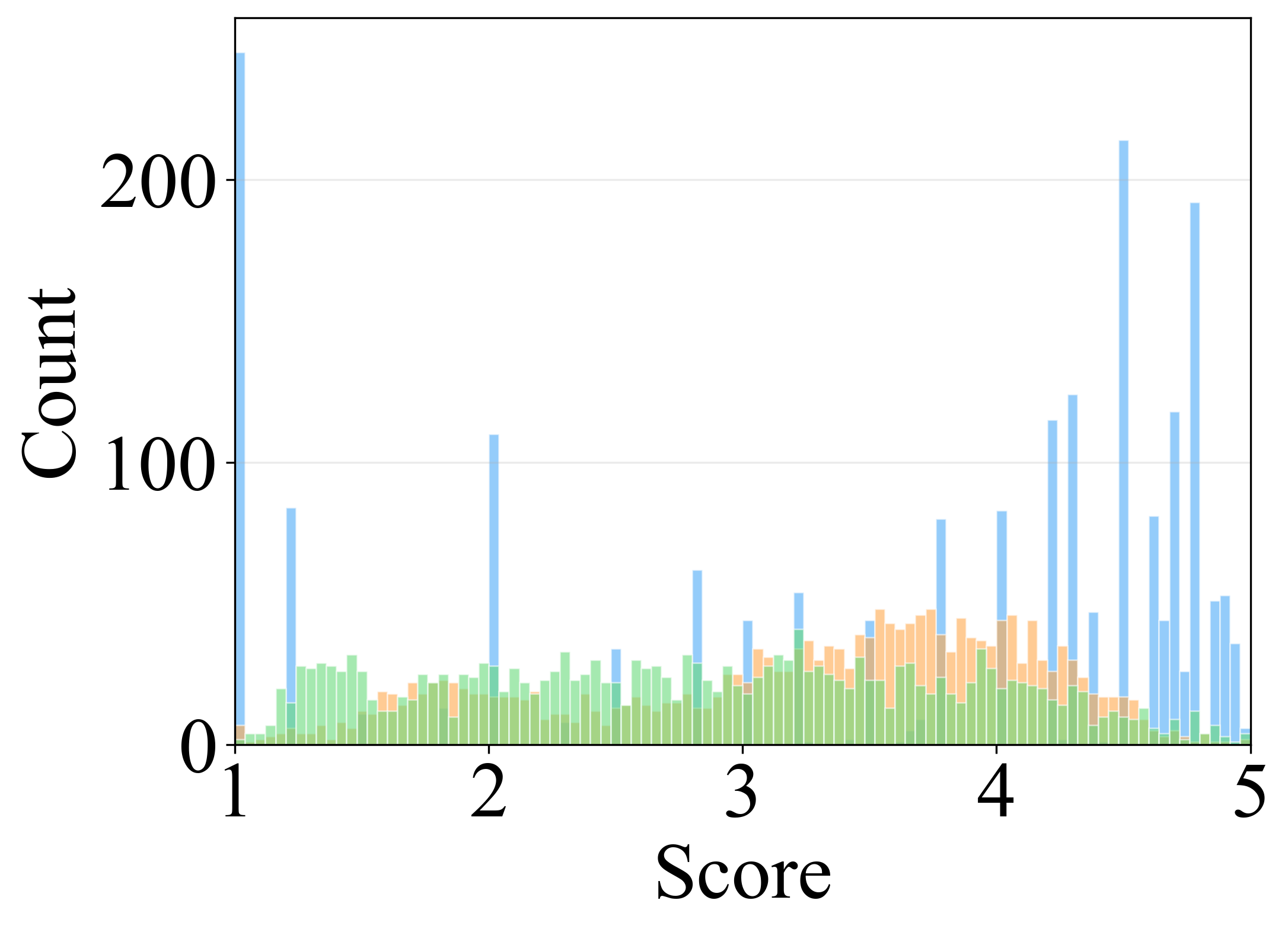}
    \subcaption{AGIQA}
    \label{fig:vqr1-agiqa}
  \end{subfigure}\hfill
  \begin{subfigure}[b]{0.25\textwidth}
    \centering
    \includegraphics[width=\textwidth]{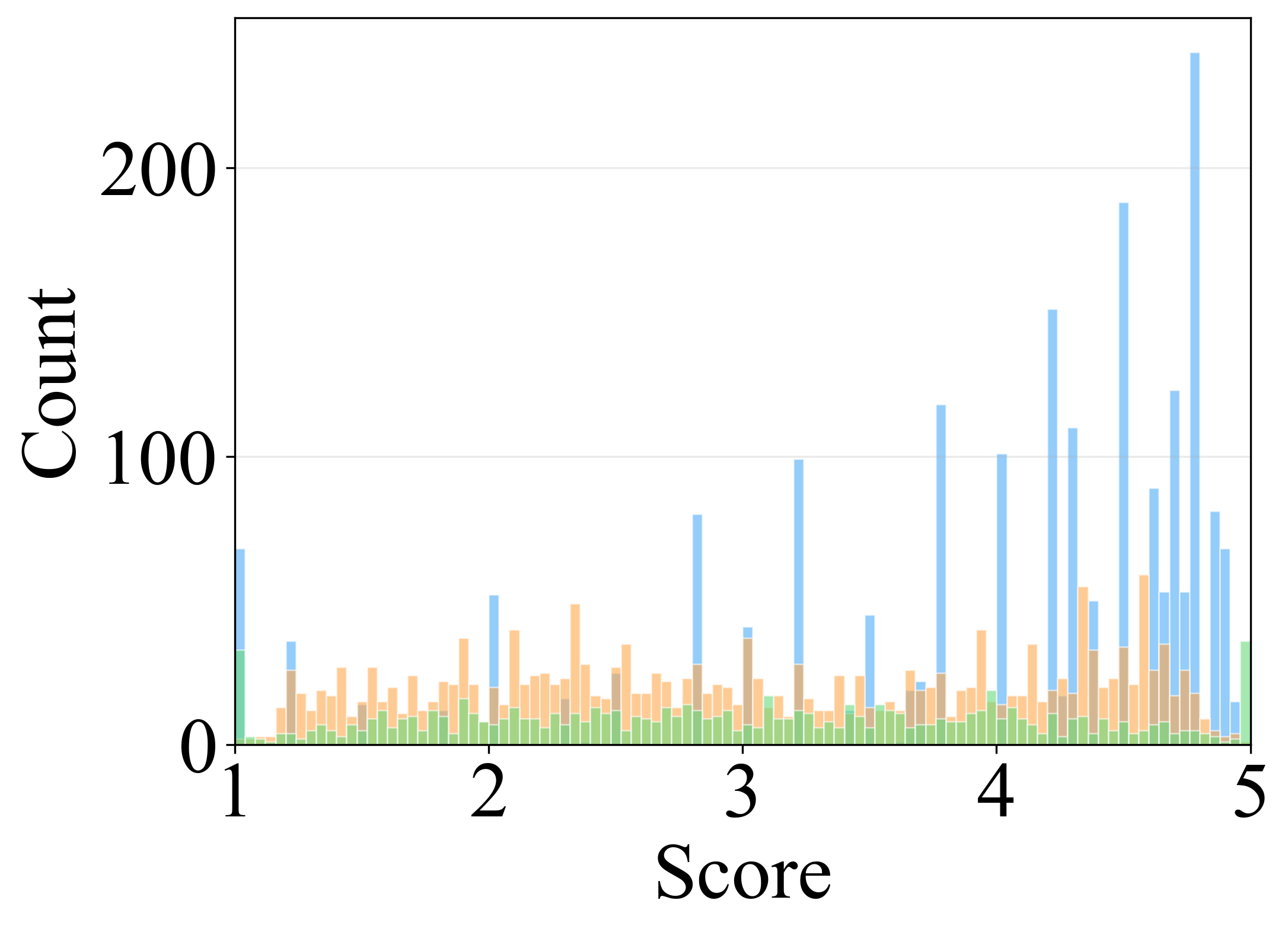}
    \subcaption{KADID}
    \label{fig:vqr1-kadid}
  \end{subfigure}\hfill
  \begin{subfigure}[b]{0.25\textwidth}
    \centering
    \includegraphics[width=\textwidth]{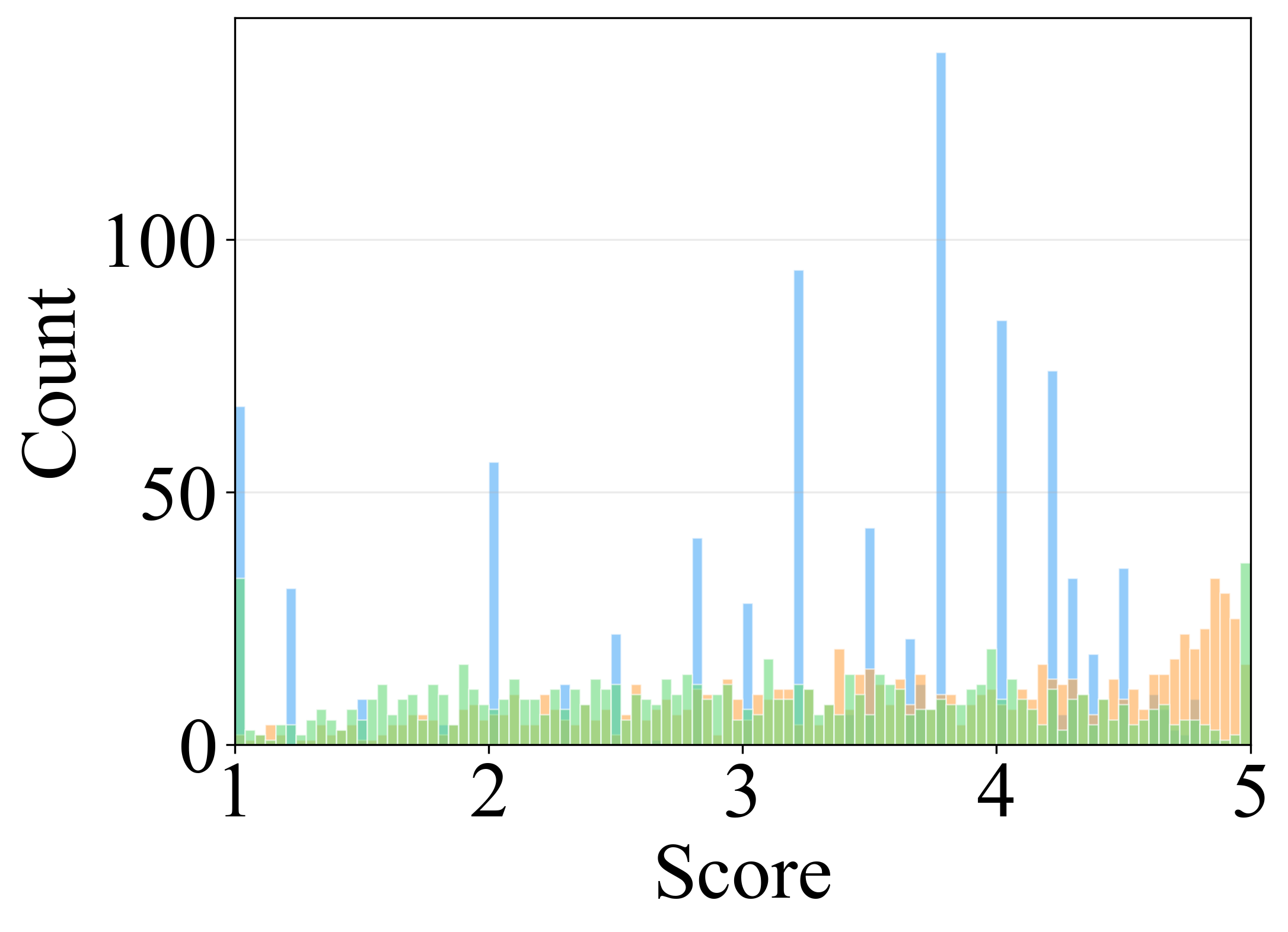}
    \subcaption{CSIQ}
    \label{fig:vqr1-csiq}
  \end{subfigure}
  \caption{Evidence and mitigation of \emph{discrete collapse} across four IQA benchmarks. Histograms compare predicted score distributions from a no-memory VLM baseline VisualQuality-R1 (blue) and ME-IQA (green) against ground-truth MOS (orange) on (a) SPAQ, (b) AGIQA, (c) KADID, and (d) CSIQ. The baseline exhibits pronounced spikes at a few discrete levels, indicative of \emph{discrete collapse}. }
  \label{fig:vqr1-dca}
\end{figure*}

\begin{table}[t]
\centering
\caption{Distributional collapse analysis across datasets. All metrics are computed with shared bins per dataset.}
\resizebox{1.0\textwidth}{!}{
\begin{tabular}{l
                ccc
                ccc
                ccc
                ccc}
\toprule
\multirow{2}{*}{\textbf{Model}}
  & \multicolumn{3}{c}{\textbf{SPAQ}}
  & \multicolumn{3}{c}{\textbf{AGIQA}}
  & \multicolumn{3}{c}{\textbf{KADID}}
  & \multicolumn{3}{c}{\textbf{CSIQ}} \\
\cmidrule(lr){2-4}\cmidrule(lr){5-7}\cmidrule(lr){8-10}\cmidrule(lr){11-13}
  & JS$\downarrow$ & Entropy$\uparrow$ & Eff.Bins$\uparrow$
  & JS$\downarrow$ & Entropy$\uparrow$ & Eff.Bins$\uparrow$
  & JS$\downarrow$ & Entropy$\uparrow$ & Eff.Bins$\uparrow$
  & JS$\downarrow$ & Entropy$\uparrow$ & Eff.Bins$\uparrow$ \\
\midrule

\multicolumn{13}{l}{\cellcolor[HTML]{EFEFEF}\textit{Q-Insight}} \\
Baseline
  & 0.49 & 2.62 & 9.10
  & 0.47 & 2.94 & 15.50
  & 0.41 & 2.98 & 16.80
  & 0.46 & 2.72 & 12.40 \\
ME-IQA
  &  0.10 &  4.23 &  58.40
  &  0.06 &  4.36 &  79.10
  &  0.06 &  4.31 &  73.60
  &  0.10 &  4.33 &  70.10 \\
\hline
  \multicolumn{13}{l}{\cellcolor[HTML]{EFEFEF}\textit{VisualQuality-R1}} \\
Baseline
  & 0.46 & 2.72 & 10.00
  & 0.46 & 3.01 & 16.14
  & 0.39 & 3.08 & 17.73
  & 0.44 & 2.83 & 13.09 \\
ME-IQA
  &  0.08 &  4.36 &  61.71
  &  0.05 &  4.49 &  84.20
  &  0.06 &  4.44 &  77.50
  &  0.09 &  4.46 &  73.95 \\
\bottomrule
\end{tabular}}
\label{tab:dist-metrics-multi}
\end{table}

\subsection{Results}
\label{sec:exp:results}

\noindent\textbf{Improvements over VLM baselines.} As summarized in Table~\ref{tab:compare_baseline}, ME-IQA consistently and substantially improves all reasoning-induced VLM baselines across diverse benchmarks. Three patterns recur: (i) gains hold across distortion regimes (authentic, AI-generated, and synthetic), suggesting that the hybrid memory serves as a cross‑domain scaffold rather than overfitting to any single corpus; (ii) improvements are larger on synthetic datasets (\eg, KADID, PIPAL), indicating that injecting local ordinal evidence sharpens sensitivity to subtle quality differences; and (iii) SRCC gains typically match or exceed PLCC gains, implying that ME-IQA primarily repairs ordinal structure without destabilizing the baseline’s global scale. These observations align with our design: reasoning‑aware retrieval provides a fine‑grained quality neighborhood, and re-ranking turns the VLM from a coarse scorer into a reliable ranker.

\noindent\textbf{Comparison with existing non-reasoning methods.} Table~\ref{tab:compare_baseline} compares ME-IQA with representative non-reasoning NR-IQA models, including: NIQE~\cite{mittal2012making}, NIMA~\cite{talebi2018nima}, MUSIQ~\cite{ke2021musiq}, CLIP-IQA+~\cite{wang2023exploring}, MANIQA~\cite{yang2022maniqa}, Q-Align~\cite{wu2024qalign}, DeQA-Score~\cite{you2025teaching}, and Compare2Score~\cite{zhu2024adaptive} (mPLUG-Owl2~\cite{ye2024mplug} as the backbone). Overall, non-reasoning methods exhibit a clear performance gap under our cross-domain evaluation protocol, with strong models such as CLIP-IQA+ and Q-Align far behind. Compare2Score is the strongest non-reasoning competitor, benefiting from comparative prompting, yet it still falls short of ME-IQA. These results indicate that (i) reasoning-induced VLMs provide richer perceptual evidence than non-reasoning scorers, and (ii) ME-IQA effectively converts such evidence into stable, dense scalar predictions at test time. 

\noindent\textbf{Test-time scaling alternatives.} Table~\ref{tab:compare_scaling} contrasts ME‑IQA with common test‑time scaling strategies, majority voting or mean aggregation over multiple samples (Maj@64, Mean@64), and Compare2Score@32~\cite{zhu2024adaptive}, under comparable compute budgets, that is 32 pairwise comparisons for ME-IQA/Compare2Score vs.\ 64 samples for Maj/Mean. We observe that: (i) ME-IQA matches or surpasses these baselines on nearly all datasets and backbones, yielding the highest WAVG SRCC/PLCC; (ii) relative to Compare2Score, our adaptive neighborhood retrieval and hybrid memory reduce anchor sensitivity, improving generalization in the presence of shift; (iii) gaps widen on difficult, fine‑grained regimes (KADID, PIPAL, CSIQ), where rank‑based refinement is most impactful; and (iv) while Mean/Maj remain competitive on authentic sets (SPAQ, LIVEW), ME-IQA still edges them out, offering a better efficiency-effectiveness trade‑off at the same FLOP budget. We also report end-to-end latency (sec/img) on $1,000$ images at 1K resolution using an NVIDIA A100 GPU with batch size $128$.
As shown in Table~\ref{tab:compare_scaling}, ME-IQA@32 has a cost comparable to Compare2Score@32, yet yields a clear accuracy gain.
Moreover, ME-IQA is 2.4$\times$ faster and more accurate than Mean@64/Maj@64.

\noindent\textbf{\emph{Discrete collapse} analysis.} As shown in Fig.~\ref{fig:vqr1-dca}, the baseline VisualQuality-R1 exhibits an obvious \emph{discrete collapse}: the score distributions concentrate at a few discrete levels. In comparison, \textsc{ME-IQA} redistributes probability mass from the spikes into the inter-level valleys, leading to denser, MOS-like profiles. To quantify this phenomenon, we report distribution statistics computed under the same binning scheme for MOS and model outputs (Table~\ref{tab:dist-metrics-multi}). First, the Jensen-Shannon divergence~\cite{nielsen2021variational} (JS) between the prediction and MOS histograms decreases (lower is better). Second, the entropy of the prediction histogram increases, and so does its entropy-derived effective number of bins (higher is better), jointly indicating broader support and closer alignment to MOS. We attribute these distributional gains to reasoning-aware neighborhood retrieval that better aligns with distortion semantics and the VLM-as-comparator that supplies informative pairwise preferences, which converts local ordinal evidence into stable, fine-grained scalar predictions. Together, these components mitigate discrete collapse while improving human-consistent calibration of the score distribution.

\begin{table}[t]
\centering
\scriptsize
\caption{Order robustness on four IQA benchmarks with 5 permutations.}
\begin{tabular}{lcc}
\toprule
\multicolumn{1}{l}{\textbf{Dataset}} & \textbf{PLCC } & \textbf{SRCC }  \\
\midrule
\multicolumn{1}{l}{SPAQ}  & 0.913$\pm$0.004 & 0.897$\pm$0.005  \\
\multicolumn{1}{l}{AGIQA} & 0.849$\pm$0.007 & 0.789$\pm$0.007  \\
\multicolumn{1}{l}{KADID} & 0.737$\pm$0.007 & 0.754$\pm$0.008  \\
\multicolumn{1}{l}{CSIQ}  & 0.783$\pm$0.006 & 0.734$\pm$0.006\\
\bottomrule
\end{tabular}
\label{tab:order_robust}
\end{table}

\begin{table}[t]
  \centering
  \scriptsize
  \caption{Effects of Neighborhood Size ($K$) on WAVG performance across seven IQA benchmarks. Relative cost is normalized to $K=32$ (\ie, $\text{Rel.\ Cost}=K/32$). 
  }
  \begin{tabular}{cccc}
    \toprule
   
    $K$ & \textbf{Rel.\ Cost} & \textbf{PLCC} & \textbf{SRCC} \\
    \midrule
    0  & 0 &   0.698    &      0.661   \\
     8   & 0.25 &   0.705    &      0.673   \\
     16  & 0.50 &    0.714   &      0.689    \\
     32{\dag} & 1.00 &   0.726    &   0.696   \\
     64  & 2.00 &   0.732    &    0.703    \\
     128 & 4.00 &    0.734   &    \textbf{0.707}    \\
     256 & 8.00 &    \textbf{0.735}   &    0.706    \\
    \bottomrule
  \end{tabular}
 \label{tab:neigh_size}
\end{table}

\begin{table}[t]
\centering
\scriptsize
\caption{Efficiency comparison. DS-$L$ denotes direct scoring with average thinking length $L$. Off. is batched offline latency, while On. and Mem. denote single-query online latency and peak memory.}
\label{tab:efficiency_ablation}
\setlength{\tabcolsep}{2.5pt}
\begin{tabular}{lccccc}
\toprule
Method & Off. $\downarrow$ & On. $\downarrow$ & Mem. $\downarrow$ & PLCC & SRCC \\
\midrule
DS-$L{=}100$ & \textbf{0.51} & \textbf{3.12} & \textbf{17} & 0.698 & 0.661 \\
DS-$L{=}300$ & 1.03 & 8.07 & 22 & 0.701 & 0.663 \\
DS-$L{=}500$ & 1.64 & 13.23 & 28 & 0.689 & 0.649 \\
ME-IQA@$8$ & 3.07 & 6.32 & 27 & 0.705 & 0.673 \\
ME-IQA@$32$ & 14.30 & 16.24 & 60 & \textbf{0.726} & \textbf{0.696} \\
\bottomrule
\end{tabular}
\end{table}

\begin{table}[t]
\centering
\scriptsize
\caption{Scoring mechanism comparison. PA is pairwise ordering accuracy on $1000$ same-content KADID pairs with MOS gap $0.1<\Delta<0.25$.}
\label{tab:scoring_mechanism}
\setlength{\tabcolsep}{4pt}
\begin{tabular}{lccc}
\toprule
Method & PLCC & SRCC & PA $\uparrow$ \\
\midrule
DS-$100$ & 0.698 & 0.661 & 57\% \\
TokenProb-Avg. & 0.682 & 0.647 & 63\% \\
RandPair-Thurstone@$32$ & 0.691 & 0.664 & 67\% \\
ME-IQA@$32$ & \textbf{0.726} & \textbf{0.696} & \textbf{75\%} \\
\bottomrule
\end{tabular}
\end{table}

\begin{table}[t]
  \centering
  \scriptsize
  \caption{Compact hyperparameter sensitivity. Extra denotes the reflection rate for $\epsilon$.}
  \label{tab:sensitivity_ablation}
  \setlength{\tabcolsep}{2.7pt}
  \renewcommand{\arraystretch}{0.92}
  \begin{tabular}{llccc}
  \toprule
  Factor & Setting & PLCC & SRCC & Extra $\downarrow$ \\
  \midrule
   & $32{:}0$ & 0.714 & 0.682 & -- \\
   & $24{:}8$ & 0.719 & 0.685 & -- \\
  $K_A{:}K_C$ & $16{:}16$ & \textbf{0.726} & 0.696 & -- \\
   & $8{:}24$ & 0.724 & 0.692 & -- \\
   & $0{:}32$ & 0.717 & 0.685 & -- \\
  \midrule
   & $0.25$ & \textbf{0.727} & 0.696 & 64\% \\
   & $0.50$ & 0.725 & \textbf{0.698} & 35\% \\
  $\epsilon$ & $0.75$ & 0.726 & 0.696 & 31\% \\
   & $1.00$ & 0.721 & 0.694 & 19\% \\
   & $1.25$ & 0.717 & 0.691 & \textbf{17\%} \\
  \bottomrule
  \end{tabular}
\end{table}

\begin{table}[t]
  \centering
  \scriptsize
  \caption{Ablations on Retrieval Embeddings. We compare (i) Random Retrieval as a baseline, (ii) Image as Embeddings, and (iii) Reasoning as Embeddings. We further include (iv) Reasoning as Embeddings* with an additional consistency reward during the RL finetuning. All runs use $K=32$ and report WAVG over seven datasets (backbone: VisualQuality-R1).}
  \begin{tabular}{lcc}
    \toprule
    \textbf{Retrieval Strategy} & \textbf{PLCC} & \textbf{SRCC} \\
    \midrule
    Random Retrieval &    0.703    &     0.670   \\
    Image as Embeddings        &    0.706    &   0.677    \\
    Reasoning as Embeddings     & 0.726  & 0.696  \\
    Reasoning as Embeddings* &   \textbf{0.737}     &    \textbf{0.710}   \\
    \bottomrule
  \end{tabular}
  \label{tab:retrieval_emb_ablation}
\end{table}

\begin{table}[t]
\centering
\scriptsize
\caption{Comparison under training/testing regimes in Compare2Score~\cite{zhu2024adaptive}.}
\begin{tabular}{l c c c c c c}
\toprule
Method & LIVE & CSIQ & KADID & BID & CLIVE & KONIQ \\
\midrule
\multicolumn{7}{l}{\cellcolor[HTML]{EFEFEF}\textit{mPLUG-Owl2}} \\
Baseline &0.902&0.899&0.903&0.887&0.861&0.910 \\
Compare2Score & 0.970 & 0.932 & 0.935 & \textbf{0.947} & 0.917 & 0.945\\
ME-IQA        & \textbf{0.973} & \textbf{0.936} & \textbf{0.943} & 0.939 & \textbf{0.925} & \textbf{0.947}\\
\midrule
\multicolumn{7}{l}{\cellcolor[HTML]{EFEFEF}\textit{VisualQuality-R1}} \\
Baseline &0.956&0.910&0.909&0.915&0.903& 0.921\\
Compare2Score &  0.953&  0.914& 0.900 & 0.912 &  0.890&  0.922\\
ME-IQA        &  \textbf{0.975}&  \textbf{0.929}& \textbf{0.936} &  \textbf{0.943}& \textbf{0.915}&  \textbf{0.955}\\
\bottomrule
\end{tabular}
\label{tab:backbone}
\end{table}

\begin{table}[t]
  \centering
  \scriptsize
  \caption{Re-ranking vs. re-scoring under the same retrieval budget ($K=32$). Metrics are computed on KADID with VisualQuality-R1.}
  \label{tab:rerank_rescore}
  \begin{tabular}{lcc}
    \toprule
    Method & PLCC & SRCC \\
    \midrule
    Re-scoring & 0.707 & 0.704 \\
    Re-ranking & \textbf{0.741} & \textbf{0.753} \\
    \bottomrule
  \end{tabular}
\end{table}

\begin{table}[t]
\centering
\scriptsize
\caption{Resolution ablation on SPAQ. ME-IQA remains stable across input resolutions.}
\label{tab:resolution_main}
\begin{tabular}{lcc}
\toprule
Resolution & PLCC & SRCC \\
\midrule
0.5K & 0.912 & 0.899 \\
1K   & 0.917 & 0.906 \\
2K   & 0.915 & \textbf{0.908} \\
4K   & \textbf{0.921} & 0.906 \\
\bottomrule
\end{tabular}
\end{table}

\begin{table}[t]
  \centering
  \scriptsize
  \caption{Effects of prior hyperparameter $\lambda$ in Thurstone fusion. All runs use $K=32$ and report results on the KADID dataset (backbone: VisualQuality-R1).}
  \label{tab:prior_lambda}
  \begin{tabular}{lccccc}
    \toprule
    \multirow{2}{*}{\textbf{Metric}} & \multicolumn{5}{c}{$\lambda$} \\
    \cmidrule(lr){2-6}
     & 0 & 1e-3 & 1e-2{\dag} & 5e-2 & 1e-1 \\
    \midrule
    PLCC & 0.724 & 0.733 & \textbf{0.741} & 0.740 & 0.737 \\
    SRCC & 0.727 & 0.739 & 0.753 & \textbf{0.755} & 0.749 \\
    \bottomrule
  \end{tabular}
\end{table}

\subsection{Ablations}
\label{sec:ablations}

Unless otherwise noted, ablations use VisualQuality‑R1 as the backbone, our default retrieval budget of $K{=}32$ with $K_{\text{A}}{=}16$/$K_{\text{C}}{=}16$, and the standard prompts and mapping described in Sec.~\ref{sec:reiqa}. We report PLCC/SRCC and WAVG (where applicable) across datasets.

\noindent\textbf{Order robustness.} To gauge sensitivity to stream order in the online setting, we evaluate ME-IQA under five random permutations of the same test split on four benchmarks (SPAQ, AGIQA, KADID, CSIQ). For each dataset we report mean $\pm$ std over permutations (Table~\ref{tab:order_robust}). Fluctuations are small and markedly below the gains that ME-IQA delivers over its baselines, indicating robust performance under varying arrival orders. We attribute this stability to (i) GT‑stratified retrieval from AM, which supplies a consistent global scaffold across the quality range, and (ii) the weak quadratic prior that regularizes refinement around the initial scale.

\noindent\textbf{Neighborhood size and efficiency.} We vary the neighborhood size $K$ and normalize relative cost to $K{=}32$ (Table~\ref{tab:neigh_size}). Accuracy improves from $K{=}0$ and saturates after $K{=}128$, with $K{=}32$ capturing most gains. Table~\ref{tab:efficiency_ablation} further shows that longer direct reasoning is not an effective substitute: DS-$L{=}300$ gives tiny gains, DS-$L{=}500$ is slower and less accurate than DS-$L{=}100$, while ME-IQA@$8$ improves over all direct baselines with comparable online latency to DS-$L{=}300$ and ME-IQA@$32$ gives the best accuracy.

\noindent\textbf{Scoring mechanism.} Table~\ref{tab:scoring_mechanism} separates ME-IQA from two alternative densification strategies. TokenProb-Avg raises PA but lowers correlations, and RandPair-Thurstone@$32$ uses the same comparison budget but remains weaker. ME-IQA reaches 75\% PA and the best correlations, indicating that retrieved local ordinal evidence is more useful than token-probability averaging or random pairwise aggregation.

\noindent\textbf{Re-ranking vs.\ re-scoring.} We compare two inference schemes under the same retrieval budget on KADID (Table~\ref{tab:rerank_rescore}). Re-scoring performs kernel regression over retrieved neighbor scores using similarity weights and therefore only interpolates within the existing calibrated score scale. This strategy improves over the direct baseline but remains substantially behind re-ranking. In contrast, ME-IQA elicits pairwise preference probabilities and fits the query score through Thurstone fusion, which introduces explicit local ordinal constraints. The gap shows that correcting \emph{discrete collapse} requires more than smoothing neighbor scores: pairwise comparisons provide richer evidence about fine-grained relative quality and yield stronger rank alignment.

\noindent\textbf{Hybrid memory and sensitivity.} Table~\ref{tab:sensitivity_ablation} shows that a balanced AM/CM split performs best, while nearby ratios remain close. The reflection gate mainly controls update frequency: performance is stable for $\epsilon{=}0.25$--$0.75$, while the reflection rate drops from 64\% to 31\%. We also vary the AM source among KONIQ, KADID, and SPAQ and test on held-out out-of-distribution datasets; the largest and average PLCC spreads are only 0.019 and 0.010, indicating stable generalization and no obvious source bias.

\noindent\textbf{Causality and cold start.} The hybrid design also addresses the cold-start issue of online IQA. At the beginning of a stream, CM is empty, so AM provides broad quality coverage and anchors the score scale before any target-domain case has been observed. As queries arrive, only past processed samples can enter CM, and the reflection gate filters updates before they are stored. Thus ME-IQA never uses future test samples or target MOS, preserving the causal online protocol. The five-parameter mapping used for AM is only a scale-alignment step between VLM scores, anchor MOS, and Thurstone fusion, rather than supervision from the target test domain. In practice, AM supplies stable global calibration while CM gradually specializes retrieval to the observed stream.

\noindent\textbf{Retrieval keys: reasoning vs.\ visuals.} Under a fixed budget, we compare retrieval keys for neighborhood construction (Table~\ref{tab:retrieval_emb_ablation}). Random Retrieval offers only modest gains; Image‑as‑Embeddings improves slightly, suggesting limited alignment between raw visual proximity and perceived quality. In contrast, Reasoning‑as‑Embeddings substantially improves retrieval fidelity and downstream re‑ranking, supporting our premise that reasoning‑grounded representations capture distortion semantics and ordinal relations. Adding a consistency reward during RL fine‑tuning to stabilize the reasoning space yields the strongest results (Reasoning‑as‑Embeddings*; see supplementary), indicating that intra‑task consistency further improves retrieval geometry.

\noindent\textbf{Resolution robustness.} ME-IQA does not rely on a fixed resizing rule. We process images at native resolution whenever feasible and otherwise follow the backbone VLM's default resizing/tokenization policy. Table~\ref{tab:resolution_main} shows that performance on SPAQ is stable from 0.5K to 4K resolution, with PLCC/SRCC varying only mildly. This suggests that the memory-enhanced comparison stage is not tied to a particular pixel scale; the retrieved reasoning summaries and pairwise preferences remain effective as long as the backbone preserves the relevant quality cues.

\noindent\textbf{Robustness to training/testing regimes.}
Table~\ref{tab:backbone} tests ME-IQA under the Compare2Score~\cite{zhu2024adaptive} protocol with a non-reasoning mPLUG-Owl2 backbone and a stronger VisualQuality-R1 backbone trained on the same six-dataset mixture. ME-IQA consistently improves both baselines and remains competitive with, or superior to, Compare2Score; the smaller mPLUG margin likely comes from using less quality-aligned vision-encoder retrieval keys.

\noindent\textbf{Prior weight in Thurstone fusion.} We sweep the quadratic prior weight $\lambda$ in the Case V objective on KADID (Table~\ref{tab:prior_lambda}). Removing the prior degrades correlations, suggesting that purely pairwise evidence can be noisy and prone to over‑correction. A very weak prior stabilizes optimization and consistently improves performance. Results peak around $\lambda\in$[1e-2,5e-2], which anchors refinement to the initial scale while preserving informative ordinal signals. Larger $\lambda$ (\eg, 1e-1) slightly harms performance, implying under‑adjustment when the prior dominates.

\noindent\textbf{Failure cases.} By aggregating neighbors and using the initial score as a weak prior, ME-IQA rarely overturns a correct direct order: on the $1000$ KADID pairs in Table~\ref{tab:scoring_mechanism}, this happens only 5 times. Remaining failures often involve subtle impulse noise or JPEG compression artifacts, suggesting a shared perceptual limitation of the underlying VLM.

\section{Conclusion}

We presented ME-IQA, a plug-and-play test-time memory-enhanced re-ranking framework that mitigates \emph{discrete collapse} in reasoning-induced VLMs. By retrieving aligned exemplars, eliciting pairwise preferences, and applying Thurstone fusion, ME-IQA produces denser calibrated scores without retraining. Across seven benchmarks and multiple backbones, it improves over reasoning and non-reasoning IQA baselines and test-time scaling alternatives. Ablations show that gains come from image-specific ordinal evidence rather than longer reasoning, token-probability averaging, random comparisons, or neighbor-score interpolation, while bounded AM/CM keeps online deployment causal and practical.

\section*{Acknowledgements}
We thank Professor Kede Ma for his helpful suggestions and discussions. Yabin Zhang is the project leader of this work.
%
%
\bibliographystyle{splncs04}
\bibliography{main}
\end{document}